\documentclass{article} 
\usepackage{iclr2026_re-align_workshop,times}

\usepackage{amsmath,amsfonts,bm}

\def\eqref#1{equation~\ref{#1}}

\def\1{\bm{1}}

\DeclareMathAlphabet{\mathsfit}{\encodingdefault}{\sfdefault}{m}{sl}
\SetMathAlphabet{\mathsfit}{bold}{\encodingdefault}{\sfdefault}{bx}{n}

\usepackage{hyperref}
\usepackage{url}
\usepackage{svg}

\title{Context Sensitivity Improves Human-Machine Visual Alignment}

\author{
\begin{tabular}[t]{c}
Frieda Born$^{1,2,3,}$\thanks{Equal contributions as first author.}\hspace{0.5em},\hspace{0.5em}
Tom Neuh\"auser$^{1,2,}$\footnotemark[1]\hspace{0.5em},\hspace{0.5em}
Lukas Muttenthaler$^{1,4,5,6,}$\thanks{Equal contributions as second author.}\hspace{0.5em},\hspace{0.5em}
Brett D. Roads$^{7,}$\footnotemark[2]\hspace{0.5em},\\
Bernhard Spitzer$^{3,8}$,\hspace{0.5em}
Andrew K. Lampinen$^{9}$,\\
Matt Jones$^{9,10}$,\hspace{0.5em}
Klaus-Robert M\"uller$^{1,2,9,11,12}$,\hspace{0.5em}
Michael C. Mozer$^{9}$ \\[3mm]
{\normalfont\small
$^{1}$Technische Universit\"at Berlin \quad
$^{2}$BIFOLD \quad
$^{3}$Max Planck Institute for Human Development} \\[0.5mm]
{\normalfont\small
$^{4}$Aignostics \quad
$^{5}$Helmholtz Munich \quad
$^{6}$Technical University of Munich 
} \\[0.5mm]
{\normalfont\small
$^{7}$University College London \quad
$^{8}$Technische Universit\"at Dresden \quad
$^{9}$Google DeepMind
 } \\[0.5mm]
{\normalfont\small
$^{10}$University of Colorado Boulder\quad
$^{11}$Korea University \quad
$^{12}$Max Planck Institute for Informatics}
\end{tabular}
}

\iclrfinalcopy 
\begin{document}

\maketitle

\begin{abstract}
Modern machine learning models typically represent inputs as fixed points in a high-dimensional embedding space. While this approach has been proven powerful for a wide range of downstream tasks, it fundamentally differs from the way humans process information. 
Because humans are constantly adapting to their environment, they represent objects and their relationships in a highly context-sensitive manner.
To address this gap, we propose a method for context-sensitive similarity computation from neural network embeddings, applied to modeling a triplet odd-one-out task with an anchor image serving as simultaneous context.
Modeling context enables us to achieve up to a 15\% improvement in odd-one-out accuracy over a context-insensitive model.
We find that this improvement is consistent across both original and ``human-aligned'' vision foundation models.
\end{abstract}

\section{Introduction}
\label{sec:introduction}
\label{sec:introduction}
Modern neural networks typically represent images and other entities as fixed points in a high-dimensional embedding space. Representing information this way has proven useful for downstream tasks that require generalization \citep{bengio2013representation, dosovitskiy2021an}, e.g., benchmarks such as ImageNet, where modern neural networks tend to outperform humans \citep{vasudevan2022dough, beyer2020imagenet,shankar2020evaluating}. However, this paradigm, reminiscent of the platonic ideal of a one-to-one-mapping between entities and representations \citep[e.g.,][]{huh2024platonic}, rests on the assumption that context does not affect how information is best encoded.

In contrast, human perception and representation are inherently adaptive, dynamically adjusting to complex and ever-changing environments composed of semantically related objects \citep{alvarez2011representing,brady2015contextual}. This dynamic nature of real-world perception suggests that the human cognitive system is optimized not for the isolated processing of entities but for the integration of complex yet structured information that exploits long-term regularities of the environment and supports flexible responses to changing contexts. Representations adapt dynamically and on a moment-to-moment basis depending on what requires attention \citep{maljkovic1994priming},  the simultaneous context \citep[e.g.,][]{bonner2021object,tversky1977features,simonson1989choice}, decoy effects \citep{huber1982adding}, goals, \citep[e.g.,][]{molinaro2023intrinsic}, and recent experiences \citep[also known as sequential dependencies;][]{mozer2007sequential,fischer2014serial,howard2002distributed,akrami2018posterior,raviv2012how,fritsche2020attraction}. Whereas rapid adaptation to the surrounding environment and its context is fundamental to human perception and cognition, it is a property that current vision machine learning models typically lack. Yet these models are increasingly being used as \emph{autoraters}, i.e., as a proxy for costly human data collection
\citep[e.g.,][]{NEURIPS2022_rlhf,christiano2017deep,stiennon2020learning}. The growing use of such systems underscores the need for model representations that adapt to changes in context. This adaptation necessitates a conceptual understanding of the world that is human-like \citep[e.g.,][]{hebart2020revealing,cichy2019spatiotemporal,peterson2018evaluating,attarian2020transforming} and, hence, inherently contextual. An alternative approach would be to remove contextual effects from human judgments \citep[e.g.,][]{mozer2011unsupervised}.

\begin{figure}[h]
\begin{center}
\includegraphics[width=0.9\linewidth]{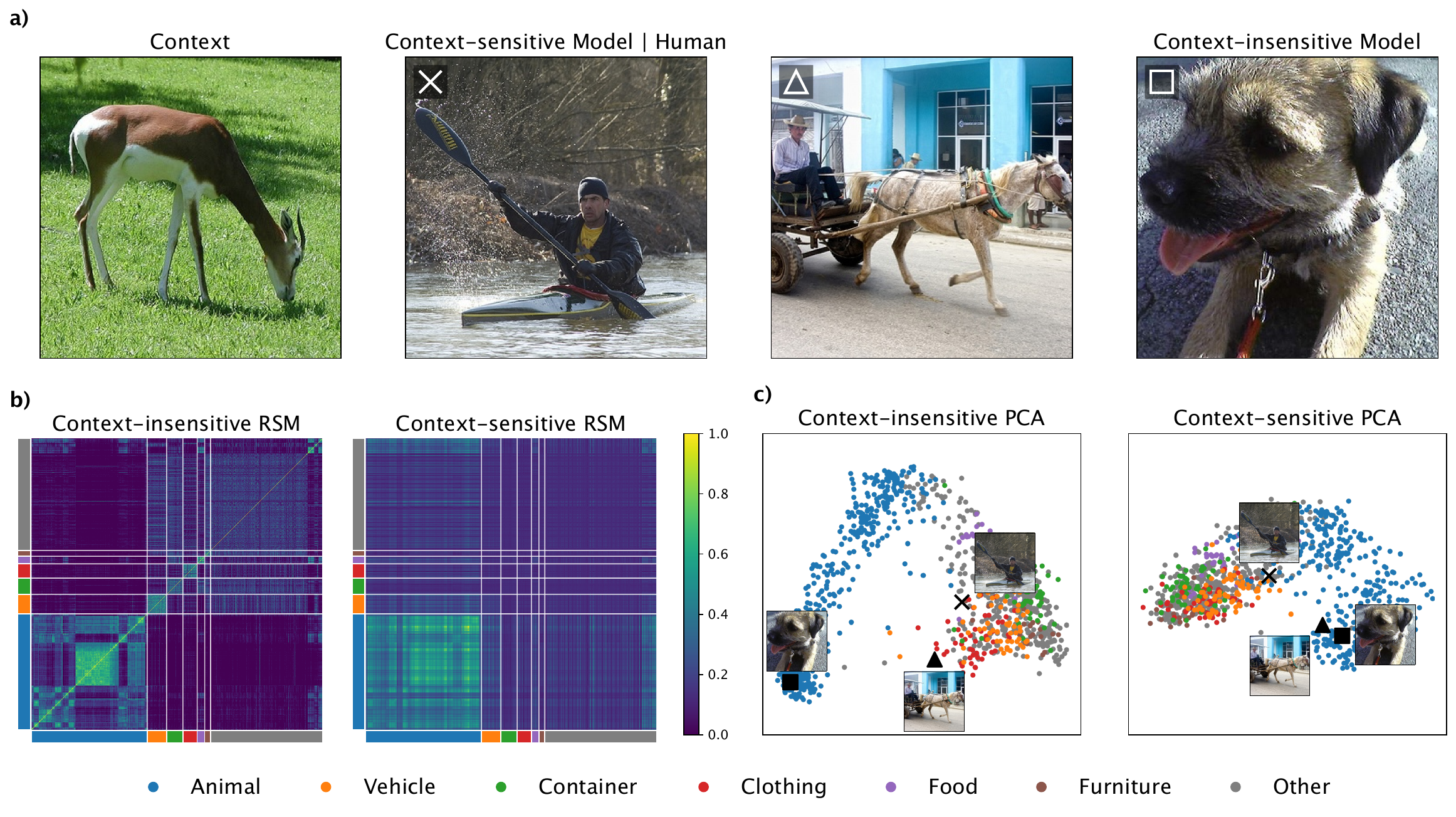}
\end{center}
\caption{Qualitative comparison of context-insensitive and context-sensitive models.
(a) Example triplet, where the human odd-one-out choice was made anchored in a given context. The context-sensitive model matches the human choice, while the context-insensitive model does not, reflecting the ambiguity of the triplet without context.
(b) Representational similarity matrices (RSMs) over a reference image set, showing how the context-sensitive model modulates similarities given the context image in a.
(c) PCA plot of the reference set and the three triplet images (same symbols as in a). The horse-cart image shifts to the animal cluster for the context-sensitive model.}
\vspace{-1em}
\label{fig:qualitative}
\end{figure}

In this work, we take a step forward in building context-sensitive models of human behavior. We utilize a large-scale behavioral dataset of human similarity judgments where participants had to judge the similarity of images with a contextual reference image \citep{roads2021enriching}. We derive predictions of human behavior from a diverse range of neural network models, by computing similarity from their embeddings. We compare models that do not consider contextual information to ones that incorporate it in a manner inspired by cognitive science theory. In addition to evaluating behavioral predictions, we examine how context influences the resulting (context-sensitive) representational spaces, showing how feature salience in decision making can shift in meaningful ways (see Fig.~\ref{fig:qualitative}). Through this work, we demonstrate how cognitively inspired interventions in machine learning systems can hold rich potential to address the need for improving the alignment between machine learning models and human representations and behavior.

\section{Related Work}
\label{sec:related_work}
\noindent{\bf Judging similarity.} Humans' ability to judge object similarity helps us organize the rich visual information surrounding us. This supports cognitive processes like object recognition, decision-making, and memory formation \citep{roads2021enriching, roads2024modeling}. Information can be compared in many ways. Thus, similarity is a flexible concept and there is no single definition of what makes things appear alike \citep{Goodman1972}. \citet{tversky1977features} famously rejected geometric theories of similarity that claim similarity is related to distance in a metric space, given evidence that human similarity judgments are asymmetric and context bound. \cite{nosofsky1986} showed that geometric similarity can provide excellent alignment with human judgment and downstream classification performance if the dimensions of the representation are rescaled with context-specific weights.

There are different experimental designs to extract human similarity ratings of concepts and study properties of human mental representation~\citep[e.g.,][]{hebart2019things, muttenthaler2022vice}, such as multi-arrangement tasks \citep{cichy2019spatiotemporal,king2019similarity}, numerical/Likert-scale pairwise similarity judgments \citep{peterson2016adapting,peterson2018evaluating}, or odd-one-out triplet tasks where subjects judge which of three items is least similar to the others \citep{robilotto2004limits,hebart2020revealing,hebart2019things,muttenthaler2022vice,fukuzawa1988internal}. Here, we use an odd-one-out triplet task setting in the reference frame of a context image \citep[][see \S\ref{sec:data} for details]{roads2021enriching}.

\noindent{\bf Human alignment.} The machine-learning community has seen growing interest in improving the alignment between human and neural network representations (e.g., \citealp{peterson2018evaluating,peterson2019human,attarian2020transforming,muttenthaler2023human,muttenthaler2023improving}; for a survey, see \citealp{sucholutsky2025getting}). Research in this area has either studied the differences between these representation spaces \citep{muttenthaler2023human, mahner2025dimensions} or aligned the representational structure of neural networks with the way humans represent information (e.g., semantically) to increase generalization and improve performance for downstream tasks, such as few-shot learning, anomaly detection, or object categorization \citep{muttenthaler2023improving, muttenthaler2025aligning, fel2022harmonizing, NEURIPS2023_dreamsim}.

\section{Methods}
\label{sec:methods}
\label{sec:data}
\noindent{\bf Data.} 
The analyses presented in this work are based on the ImageNet-HSJ dataset introduced by \citet{roads2021enriching}, which consists of large-scale human similarity judgments for images of the ImageNet ILSVRC validation set. The dataset was originally collected in an 8-choose-2 task setting, in which participants view a display of nine images in a grid, with a query image placed in the center, and select two
reference images most similar to the query image \citep{roads2021enriching}. For our primary analysis, this dataset was restructured into a triplet-with-context task format with the query image as context, an example of which is shown in Fig.~\ref{fig:qualitative}a. Each triplet trial is composed of the two reference images chosen in a given trial along with one unselected image from the same trial, resulting in 6 triplet trials derived from one 8-choose-2 trial. The task is to predict the odd-one-out \emph{in the context of the query}. For details on data preprocessing and splits, see Appx.~\ref{sec:data_preprocessing_and_splits}.

\begin{figure}[h]
\begin{center}
\includegraphics[width=0.9\linewidth]{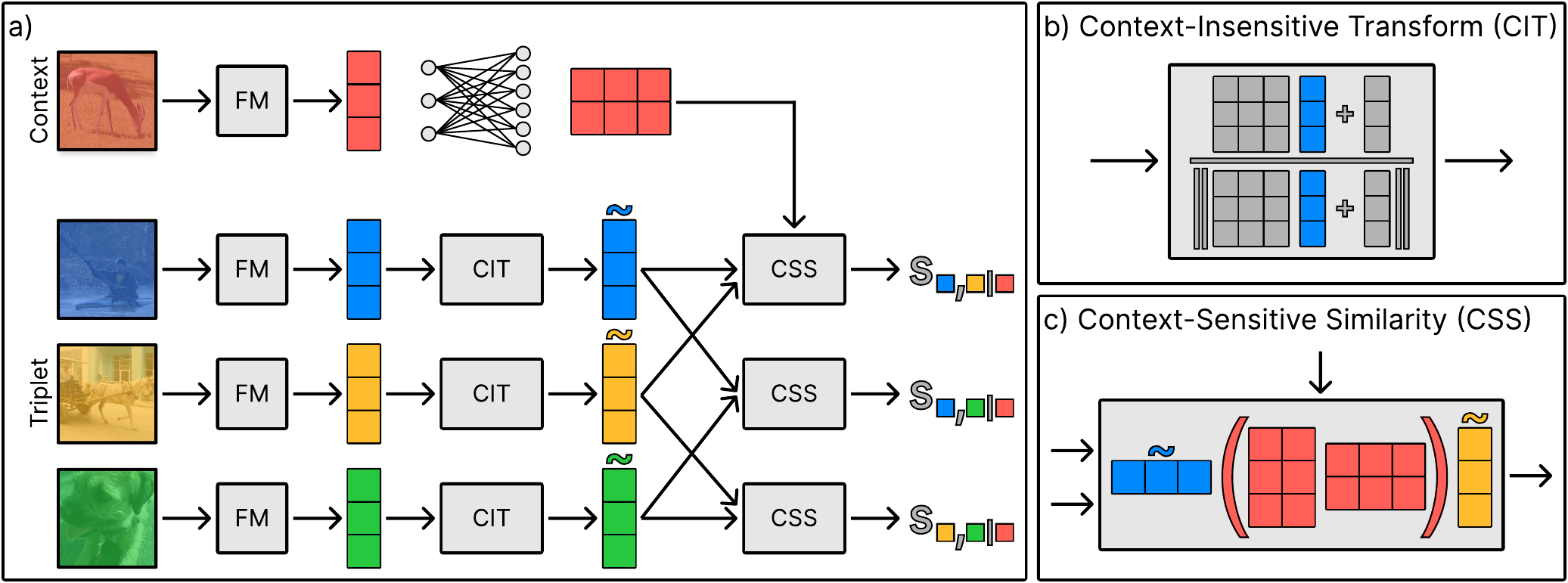}
\end{center}
\caption{Method overview. A foundation model (FM) extracts embeddings for the 
context and triplet images. The triplet embeddings undergo a context-insensitive transform (CIT), and the context embedding is mapped to 
a matrix that is used to compute pairwise, context-sensitive similarities (CSS) between the transformed triplet embeddings. The trainable parameters consist of those of the context-insensitive transform and the neural network that maps the context embedding to the matrix.}
\label{fig:method}
\end{figure}

\label{sec:model}
\noindent{\bf Model.} We propose a context-sensitive model that takes as input a triplet of images along with a context image and produces an odd-one-out decision over the triplet conditioned on the context. The model is shown in Fig.~\ref{fig:method}a. The model processes each image independently through a foundation model (FM) to obtain an embedding. A \emph{context-insensitive transform} is then applied to improve alignment with human similarity judgments for our triplet-with-context task. A neural network, consisting of a single fully connected linear layer, maps the context-image embedding to a low-rank matrix, which is then used to calculate context-dependent similarity scores between pairs of image embeddings. Having obtained this \emph{context-sensitive similarity} for each image pair in the triplet, we identify the pair with the greatest similarity and deem the remaining image to be the odd-one-out.

\noindent{\bf Feature extraction.} To embed triplet and context images, we extract features using FMs from \citet{muttenthaler2025aligning} trained with three different pretraining objectives: supervised~\citep[ViT;][]{dosovitskiy2021an}, image/text contrastive~\citep[SigLIP;][]{zhai2023sigmoid}, and self-supervised~\citep[DINOv2;][]{oquab2024dinov}. For each, \citet{muttenthaler2025aligning} provides \emph{human-aligned} versions, obtained via fine-tuning to better match human similarity judgments, learning a static transformation of the embedding space. Our approach adds an explicit context-sensitive component, enabling similarities to dynamically shift with context rather than rely on fixed  representations.

\noindent{\bf Context-insensitive transform.} Although we use FMs fine-tuned for human-alignment, this may not generalize to the triplet-with-context task. To improve alignment on our task, we apply an affine transform followed by normalization to the embedding $x_i \in \mathbb{R}^d$ of image $i$:
\begin{equation*}
\tilde{x}_i = \frac{W x_i + b}{\lVert W x_i + b\rVert},
\end{equation*}
where $W \in \mathbb{R}^{d \times d}$ and $b \in \mathbb{R}^d$ are learned parameters. Normalization is motivated by prior work indicating that the vector norm provides no useful categorical information \citep{scott2021}.

\noindent{\bf Context-sensitive similarity.}
To compute the context-sensitive similarity $s_{i,j \mid c} \in \mathbb{R}$ between images $i$ and $j$ conditioned on context image $c$, we define:
\begin{equation*}
s_{i,j \mid c} = \tilde{x}_i^\top A_{c}\,\tilde{x}_j,
\end{equation*}
where $A_c \in \mathbb{R}^{d \times d}$ is a context-dependent symmetric positive semidefinite matrix.
To enforce this constraint, we parameterize $A_{c} = B_{c}^\top B_{c}$, with $B_{c} \in \mathbb{R}^{r \times d}$ and $r \ll d$.
The matrix $B_c$ is the output of a neural network applied to the context image embedding.
This yields a context-dependent kernel that corresponds to the dot product between projected embeddings $B_c \tilde{x}_i, B_c \tilde{x}_j \in \mathbb{R}^r$.

\noindent{\bf Odd-one-out decision.}
Given a triplet of images $\{p,q,r\}$ and a context image $c$, the model chooses image $k \in \{p, q, r\}$ to be the odd-one-out according to the softmax probability for the similarity of the pair of non-chosen images $\{i,j\} = \{p,q,r\} \setminus \{k\}$, as $\Pr(k \mid \{p,q,r\}, c) \propto \exp\!\big(s_{i,j \mid c}\big)$.

\noindent{\bf Alignment loss and identity regularization.}
We adapt the approach of \citet{muttenthaler2023improving} to the context-sensitive setting. Similarly, we train the model via maximum-likelihood estimation with respect to the model parameters $\theta$, consisting of the affine transform and the neural network parameters, to predict the human odd-one-out choice. For a set of images $s$, we denote the human-chosen oddball as $k_s^{\ast}$, leading to a loss defined over $n$ training instances:
\begin{equation*}
\mathcal{L}(\theta) = -\frac{1}{n}\sum_{s=1}^n \log \Pr(k_s^{\ast} \mid \{p_s,q_s,r_s\}, c_s) + \lambda_1 \lVert W - I\rVert_F^2 + \frac{\lambda_2}{n}\sum_{s=1}^n \lVert A_{c_s}-I\rVert_F^2,
\end{equation*}
where the latter two terms preserve the pretrained representational structure. Specifically, the second term aims to keep the context-independent transformation close to the original representation and the third term attempts to keep the context-sensitive similarity close to the cosine similarity.

\section{Experimental Results}
\label{sec:experimental_results}
\begin{figure}[h]
\begin{center}
\includegraphics[width=0.9\linewidth]{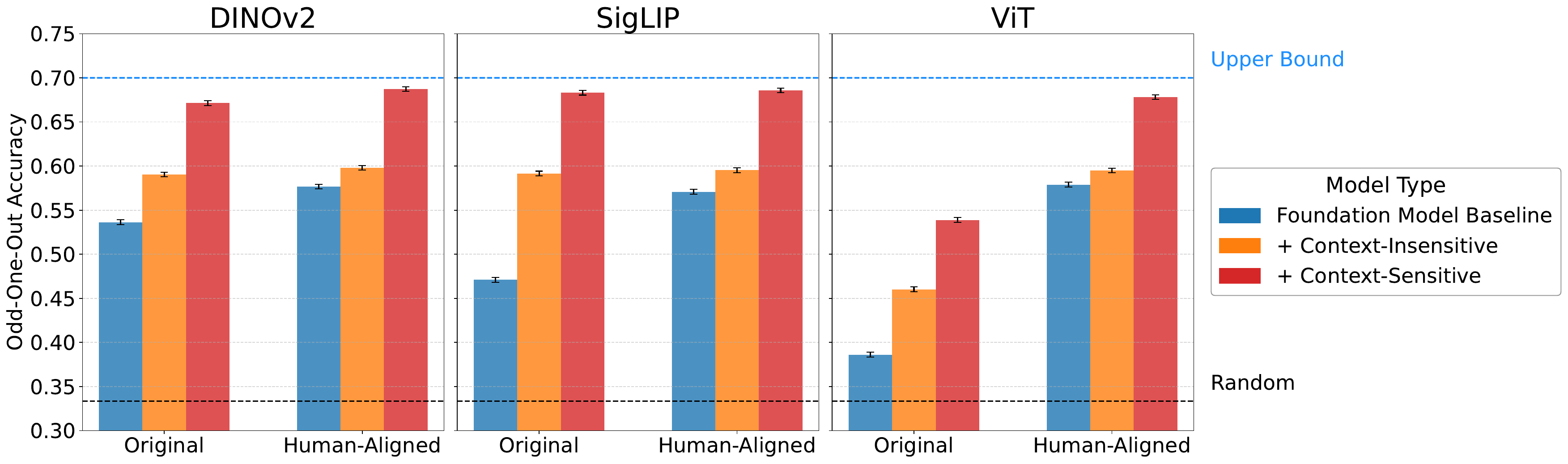}
\end{center}
\caption{Odd-one-out accuracy on our triplet-with-context task for original and human-aligned DINOv2, SigLIP, and ViT bases. The context-sensitive models consistently outperform both FM baselines and context-insensitive transforms. Error bars denote 95\% confidence intervals computed over held-out test trials. See Appx.~\ref{sec:upper_bound} on upper bound estimation.}
\vspace{-1em}
\label{fig:quantitative}
\end{figure}

\noindent{\bf Quantitative Results.} To evaluate the impact of context-sensitive modeling, we assessed odd-one-out accuracy on our triplet-with-context task.
In this analysis, we considered three differently trained FMs (i.e., DINOv2, SigLIP, ViT), in addition to their human-aligned variants obtained via AligNet fine-tuning \citep{muttenthaler2025aligning}. 
For each of the six FMs, we compare the FM baseline to our context-sensitive model and to a context-insensitive model, which removes the context-sensitive similarity computation and retains only the context-insensitive transform. Experimental details are provided in Appx.~\ref{sec:experimental_details}. Across all conditions, both the context-insensitive and context-sensitive models improved over the FM baseline, with the context-sensitive model yielding the largest gains. Paired bootstrap analyses on held-out test trials confirmed that all stepwise improvements were reliable, with 95\% confidence intervals excluding zero in every comparison (see Appx.~\ref{sec:stats} for further details). Fig.~\ref{fig:quantitative} presents six sets of bars indicating odd-one-out accuracy on a held-out test set of data (for detailed performances across models, see Appx.~\ref{sec:quantitative_table}). For each set, the FM baseline (blue bar) consistently shows a weaker odd-one-out prediction accuracy than a variant with only a context-insensitive transform (orange bar). Critically, incorporating a context-sensitive modulation of the embeddings (red bar) further improves accuracy. These stepwise gains are observed across all three FMs and both original and human-aligned representations.

\label{sec:quantitative}

\noindent{\bf Qualitative Results.} To assess the effect of context-sensitive modeling, we visualize representational similarity matrices~\citep[RSMs;][]{kriegeskorte2008representational} and PCAs of the representation spaces produced by the context-insensitive and context-sensitive models for the example triplet and context shown in Fig.~\ref{fig:qualitative}a; for more examples, see Appx.~\ref{sec:qualitative_examples}. For PCA, we interpret the projected embeddings as the context-sensitive representations. As shown in Fig.~\ref{fig:qualitative}a, the context-insensitive model selects the dog as the odd-one-out. Consistently, the context-insensitive RSM (Fig.~\ref{fig:qualitative}b, left) assigns higher similarity to the canoe and horse-cart images, both showing vehicles, than to the dog and horse-cart images, both containing animals. In the corresponding PCA plot (Fig.~\ref{fig:qualitative}c, left), the dog clusters with other animal images and is separated from the vehicle-related images, including the horse-cart. Given the context image of a deer, humans select the canoe as the odd-one-out, a choice mirrored by the context-sensitive model (Fig.~\ref{fig:qualitative}a). In the context-sensitive RSM (Fig.~\ref{fig:qualitative}b, right), similarities between images containing animals are upweighted, while similarities between images of vehicles are downweighted. In the corresponding PCA plot (Fig.~\ref{fig:qualitative}c, right), this modulation pushes the horse-cart image toward the animal cluster, reflecting its reinterpretation in the given context.
\vspace{-1em}
\label{sec:qualitative}

\section{Discussion}
\label{sec:discussion}
We introduce a psychologically plausible, context-sensitive modulation of neural network representations, and demonstrate that it improves the degree of alignment with human judgments. Our approach draws inspiration from cognitive theories of similarity~\citep[e.g., Tversky's features of similarity;][]{tversky1977features}, introducing selective feature re-weighting in the presence of contextual cues. Qualitative analyses highlight a context-induced reshaping of the embedding space. Notably, our model re-weights arbitrary directions in the embedding space whereas human dimensional attention is restricted to predetermined axes \citep{kruschke1993}.
Quantitatively, our context-sensitive modulation leads to substantial improvements in predicting human odd-one-out judgments across a range of state-of-the-art FMs. Interestingly, combining context-sensitive modulation with human-alignment via fine-tuning \citep{muttenthaler2025aligning} yields remarkably similar performance levels across different FMs. Relative benefits from both human alignment and contextual modulation seem to be largest for supervised models, suggesting that this kind of pretraining benefits the most from re-organizing the representation space reflecting context. These results indicate the potential of cognitively inspired modelling to better reflect human behavior.

Our current work has limitations (see Appx.~\ref{sec:limitations}). Future approaches ought to incorporate additional sources of contextual bias, including sequential dependencies \citep[e.g.,][]{mozer2010improving,mozer2011unsupervised,jones2013sequential,jones2022learning,akrami2018posterior,fritsche2020attraction}, internal goals or motivational states~\citep[e.g.,][]{molinaro2023intrinsic} and expand to vision–language models capable of visual reasoning across multiple images~\citep{zhao2024benchmarkingmultiimageunderstandingvision}.
Although the odd-one-out task treats the image comparisons as symmetric, other human judgment tasks require asymmetric comparison to a reference image. In such cases, Tversky's observation of asymmetries in similarity ought to come into play to further improve human-machine alignment.

\noindent{\bf Acknowledgments.} We thank Maria Eckstein for helpful feedback and insightful comments on this work. We also thank Lorenz Linhardt for helpful discussions and contributions to the conceptualization of the project. Finally, we thank Anouk Bielefeldt and Hannah Louisa Boldt for their assistance with research support tasks.

\bibliography{iclr2026_conference}
\bibliographystyle{iclr2026_conference}

\clearpage

\appendix
\section{Disclosure on the Use of Generative Artificial Intelligence}
Generative artificial intelligence was used to improve grammar and wording and to assist with writing experiment code. The submission is primarily human authored and the authors hold responsibility for all content of the submission.

\section{Limitations}
\label{sec:limitations}
One limitation of the present study concerns the behavioral dataset used for training and evaluation. In the original ImageNet-HSJ dataset introduced by \citet{roads2021enriching}, each trial was presented to only a single participant. As a result, our reconstructed dataset likewise contains one participant judgment per trial, meaning that models are trained on individual responses rather than on a distribution of judgments reflecting consensus in object similarity across participants. Although this data collection strategy is common in large-scale similarity datasets, including the THINGS dataset \citep{hebart2019things}, it may introduce additional variability in the training signal. A second limitation concerns the construction of the triplet-with-context dataset itself. Triplets were derived by pairing the two images selected as most similar to the query with each unselected image. While this reconstruction provides a practical way to derive contextual triplet comparisons and increases dataset size, it may introduce statistical regularities related to the dataset construction. In particular, the conversion implicitly hypothezises that the two selected items are more similar to each other than to unselected images, an assumption not explicitly required by the original behavioral task.

\section{Data Preprocessing and Splits}
\label{sec:data_preprocessing_and_splits}
We filtered out trials in which two or more triplet images were in the same ImageNet class, ensuring that trivial choices were excluded. For the training and evaluation of our models, we created training, validation, and test splits on the trial level (of the original 8-choose-2 task), stratified by participant, such that all participants proportionally contributed an equal amount of trials to each of the splits. The resulting splits included N = 1,003,759 trials for training, N = 125,646 trials for validation, and N = 125,710 trials for testing.

\section{Experimental Details}
\label{sec:experimental_details}

In all experiments, the base variants of the FMs were used to extract 768-dimensional representations. All context-sensitive and context-insensitive models were trained for 50 epochs using stochastic gradient descent. The learning rate and batch size were set to $0.001$ and $128$, respectively. A grid search was conducted to find the best values for the following hyperparameters:

\begin{itemize}
    \item the rank $r \in \{16, 32\}$ of the matrix $B_c$,
    \item the regularization strength $\lambda_1 \in \{0.0001, 0.001\}$ for the context-insensitive transform,
    \item the regularization strength $\lambda_2 \in \{10^{-5}, 10^{-4}, 10^{-3}\}$ for the context-sensitive similarity,
    \item and the softmax temperature $\tau \in \{1.0, 5.0, 7.5\}$.
\end{itemize}

The hyperparameter configuration with the highest validation accuracy was selected, and the corresponding model was evaluated on a held-out test set.

\section{Statistical Analysis of Model Comparisons}
\label{sec:stats}

To assess the reliability of performance differences between models, we conducted paired bootstrap analyses for all pairs of models shown in Fig.~\ref{fig:quantitative}. For each pair, we report the mean difference in accuracy ($\Delta$) along with 95\% confidence intervals (CI) computed over held-out test trials.

Across all FMs, both the context-insensitive and context-sensitive models improved over the FM baselines, and the context-sensitive models consistently show higher prediction accuracy compared to the context-insensitive models. All reported confidence intervals exclude zero, indicating reliable differences.

\paragraph{DINOv2.}
For the original representation, the context-insensitive model improved over the baseline by $\Delta = 0.054$ [0.052, 0.057], and the context-sensitive model by $\Delta = 0.135$ [0.132, 0.138]. The additional gain of the context-sensitive over the context-insensitive model was $\Delta = 0.081$ [0.078, 0.083]. 
For the human-aligned representation, improvements were $\Delta = 0.021$ [0.019, 0.023] (context-insensitive vs. baseline), $\Delta = 0.111$ [0.108, 0.114] (context-sensitive vs. baseline), and $\Delta = 0.089$ [0.087, 0.092] (context-sensitive vs. context-insensitive).

\paragraph{SigLIP.}
For the original representation, improvements were $\Delta = 0.120$ [0.117, 0.123] (context-insensitive vs. baseline), $\Delta = 0.212$ [0.209, 0.215] (context-sensitive vs. baseline), and $\Delta = 0.092$ [0.089, 0.094] (context-sensitive vs. context-insensitive). 
For the human-aligned representation, improvements were $\Delta = 0.025$ [0.022, 0.026], $\Delta = 0.115$ [0.112, 0.118], and $\Delta = 0.090$ [0.088, 0.093], respectively.

\paragraph{ViT.}
For the original representation, improvements were $\Delta = 0.074$ [0.071, 0.077] (context-insensitive vs. baseline), $\Delta = 0.153$ [0.149, 0.156] (context-sensitive vs. baseline), and $\Delta = 0.079$ [0.075, 0.082] (context-sensitive vs. context-insensitive). 
For the human-aligned representation, improvements were $\Delta = 0.016$ [0.014, 0.018], $\Delta = 0.099$ [0.097, 0.102], and $\Delta = 0.083$ [0.081, 0.086], respectively.

\section{Upper Bound Estimation}
\label{sec:upper_bound}
Each 8-choose-2 trial, and thus each triplet-with-context trial, is observed by exactly one participant and therefore yields only a single response. Estimating an upper bound on odd-one-out accuracy requires multiple responses per trial. To address this, we identify trials by their ImageNet class labels rather than image identities. As a result, multiple image-level trials are mapped to the same class-level trial, yielding multiple responses for a subset of trials.

For each trial in this subset, we estimate the empirical distribution of odd-one-out responses. Let this distribution be $p = (p_1, p_2, p_3)$, where $p_i$ denotes the probability that the $i$-th image in the triplet is selected as the oddball. The optimal strategy is to predict the image with the highest probability, yielding an expected odd-one-out accuracy of $\max_i p_i$. 

We estimate the upper bound on odd-one-out accuracy as the average of $\max_i p_i$ across the subset of trials with multiple responses.

\section{Supplementary Results}

\subsection{Quantitative Results}
\label{sec:quantitative_table}

\begin{table}[h]
\centering
\caption{Triplet odd-one-out accuracies for original and human-aligned foundation model bases on the triplet-with-context task.
We report accuracies for the foundation model baselines (FM), context-insensitive models (CI),
and context-sensitive models (CS). Relative improvements are computed with respect to the foundation
model baselines and context-insensitive models.}
\begin{tabular}{l|c|c|c|c|c}
 & FM & CI & CS & \textuparrow \ FM (\%) & \textuparrow \ CI (\%) \\
\hline
DINOv2 (original) & 0.536 & 0.590 & 0.657 & 22.4 & 11.2 \\
DINOv2 (human-aligned) & 0.577 & 0.598 & 0.687 & 19.2 & 14.9 \\
SigLIP (original) & 0.471 & 0.582 & 0.682 & 44.9 & 17.4 \\
SigLIP (human-aligned) & 0.571 & 0.595 & 0.685 & 20.0 & 15.0 \\
ViT (original) & 0.386 & 0.460 & 0.539 & 39.6 & 17.1 \\
ViT (human-aligned) & 0.579 & 0.585 & 0.672 & 16.1 & 14.8 \\
\end{tabular}
\label{tab:quantitative_results}
\end{table}

\subsection{Qualitative Results}
\label{sec:qualitative_examples}

\begin{figure}
\begin{center}
\includegraphics[width=0.95\linewidth]{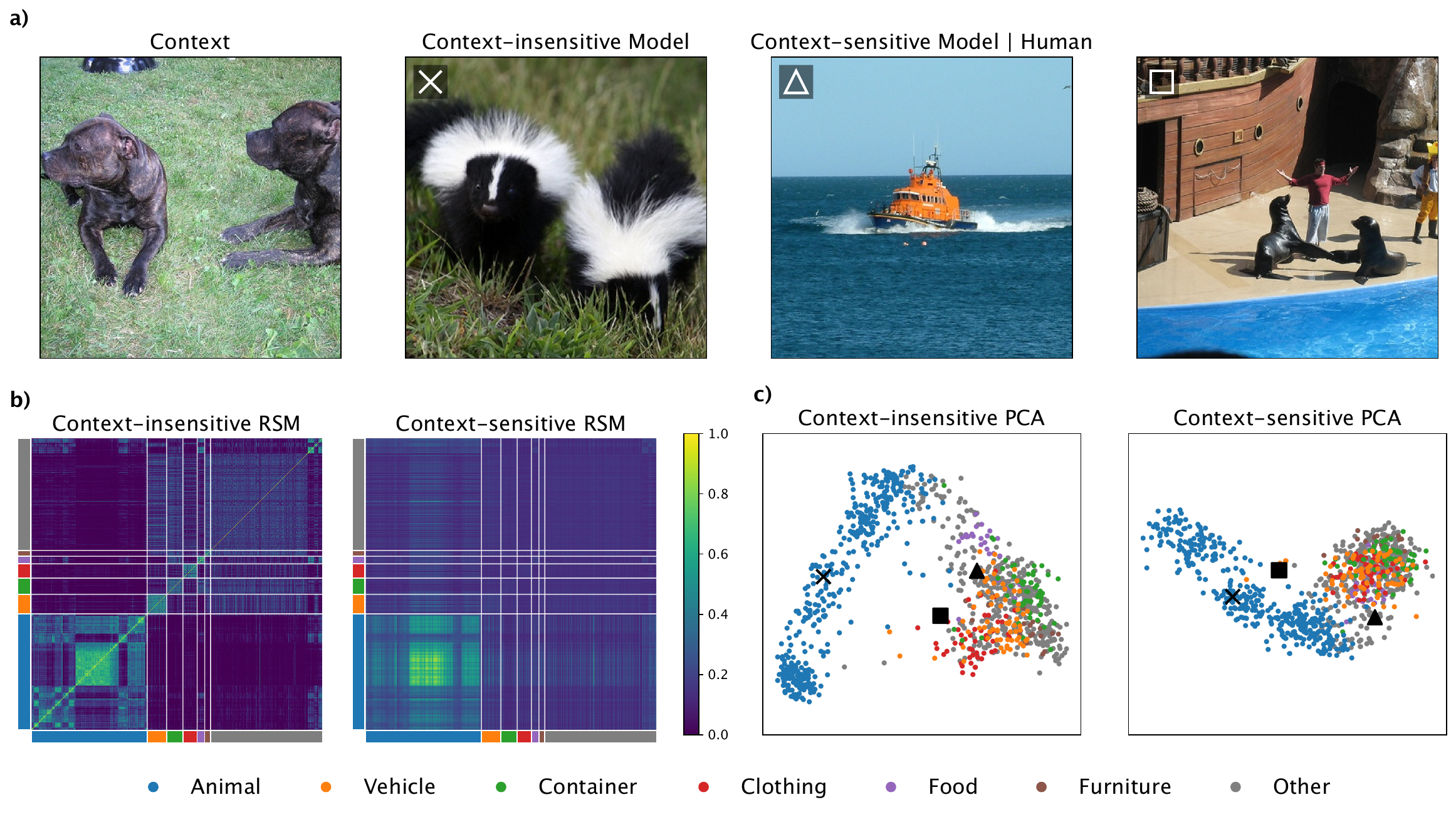}
\includegraphics[width=0.95\linewidth]{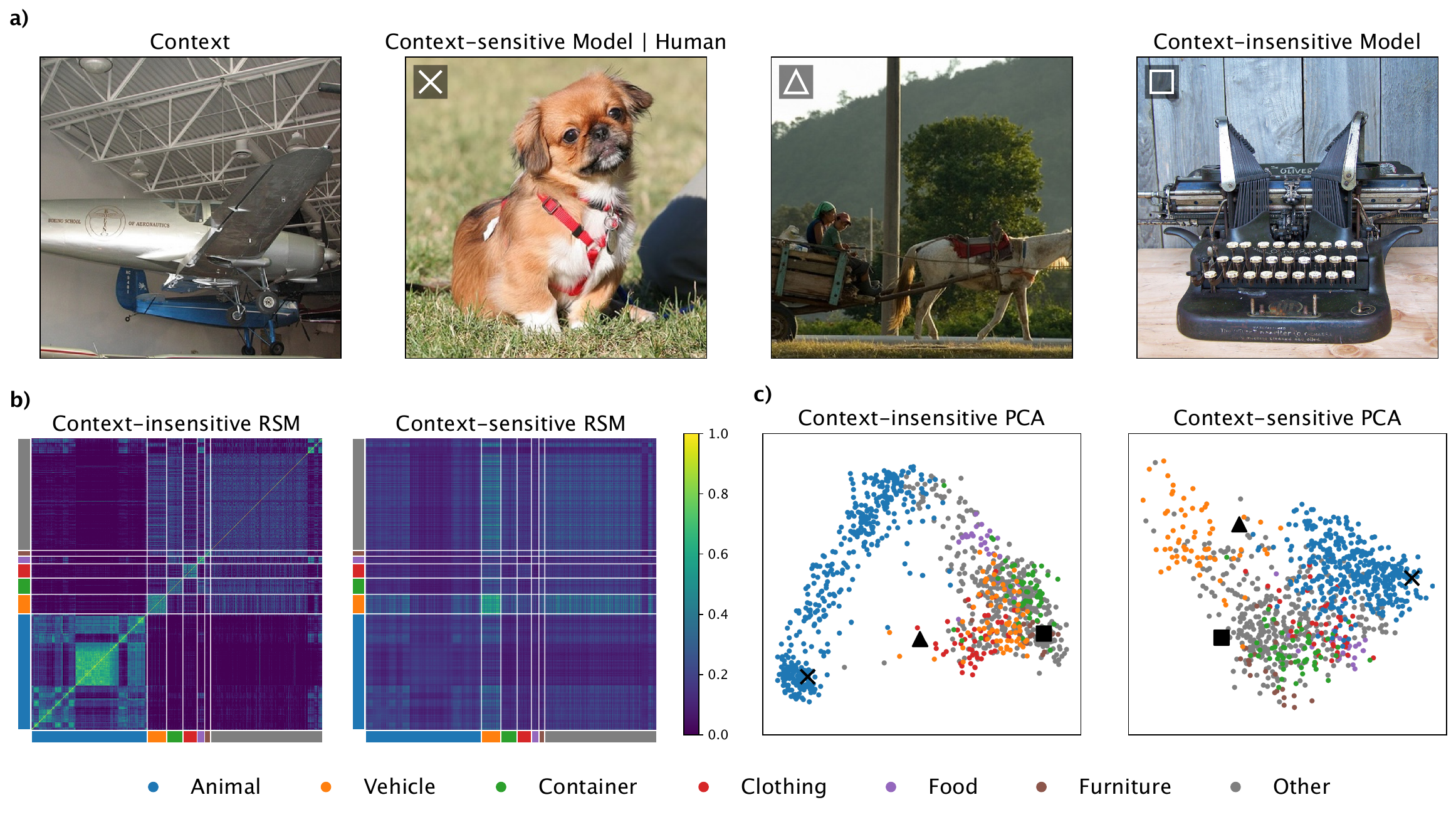}
\includegraphics[width=0.95\linewidth]{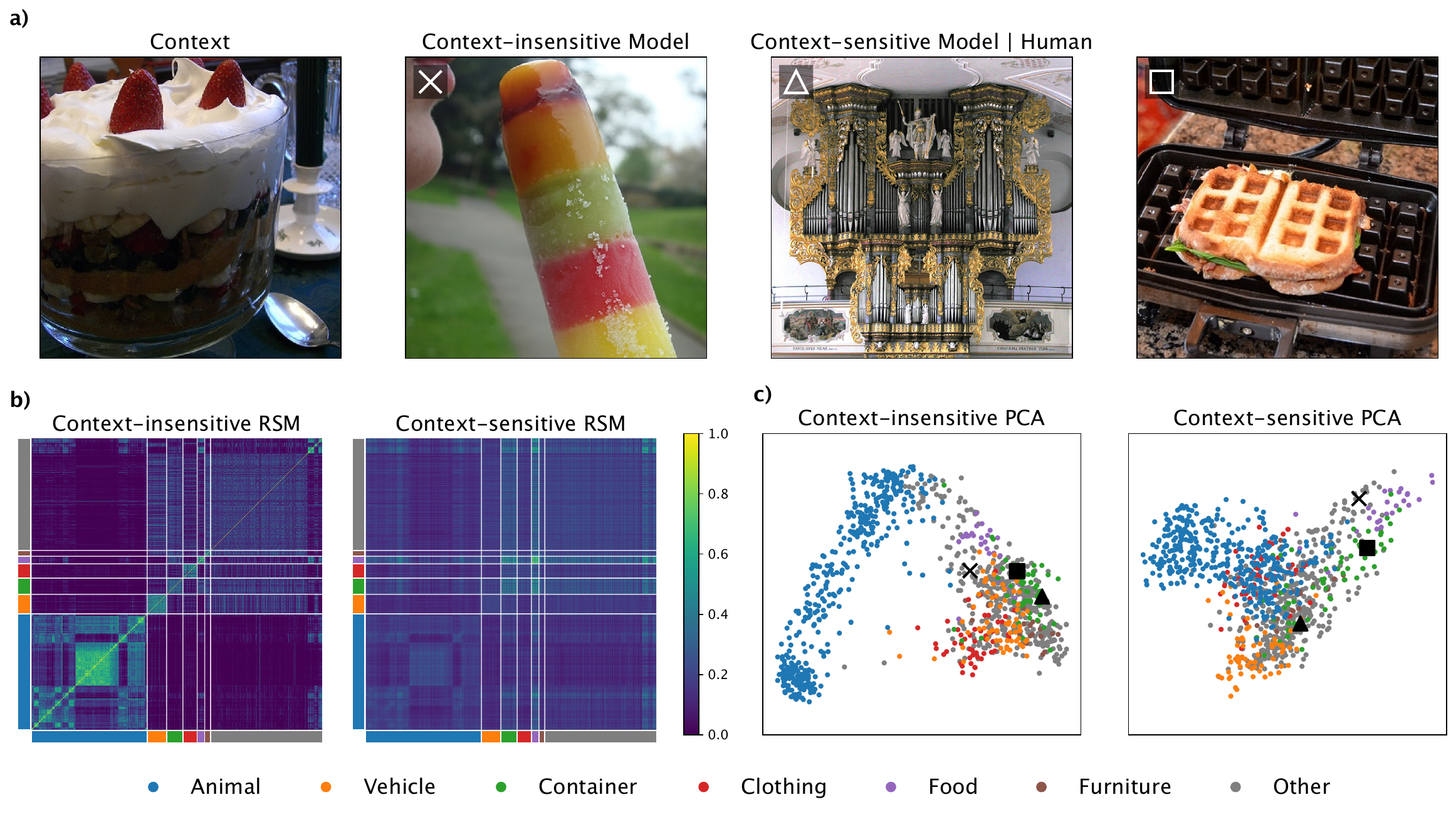}
\end{center}
\end{figure}

\clearpage

\end{document}